\documentclass[runningheads]{llncs}

\usepackage[T1]{fontenc}
\usepackage{graphicx}
\usepackage{subfigure}
\usepackage{cite}
\usepackage[colorlinks,
            linkcolor=blue,      
            anchorcolor=blue,  
            citecolor=blue,       
            ]{hyperref}
\usepackage{amsmath}
\usepackage{amssymb}
\usepackage{marvosym}

\makeatletter
\def\thanks#1{\protected@xdef\@thanks{\@thanks
        \protect\footnotetext{#1}}}
\makeatother
\begin{document}
\title{Fine-Grained Urban Flow Inference with Dynamic Multi-scale Representation Learning}

\author{Shilu Yuan\inst{1}, Dongfeng Li\inst{2}, Wei Liu\inst{3}, Xinxin Zhang\inst{1}, Meng Chen\inst{1}, Junjie Zhang\inst{4}, \and Yongshun Gong\inst{1}\textsuperscript{(\Letter)} }
\authorrunning{S.Yuan Author et al.}
%
\institute{Shandong University, Jinan, China \\
 \and University of Washingtong, Seattle, America\\
 \and University of Technology Sydney, Sydney, Australia\\ 
 \and Shanghai University, Shanghai, China }
\titlerunning{UrbanMSR: A Model For FUFI} 
\maketitle              

\begin{abstract}
Fine-grained urban flow inference (FUFI) is a crucial transportation service aimed at improving traffic efficiency and safety. FUFI can infer fine-grained urban traffic flows based solely on observed coarse-grained data. However, most of existing methods focus on the influence of single-scale static geographic information on FUFI, neglecting the interactions and dynamic information between different-scale regions within the city. Different-scale geographical features can capture redundant information from the same spatial areas. In order to effectively learn multi-scale information across time and space, we propose an effective fine-grained urban flow inference model called UrbanMSR, which uses self-supervised contrastive learning to obtain dynamic multi-scale representations of neighborhood-level and city-level geographic information, and fuses multi-scale representations to improve fine-grained accuracy. The fusion of multi-scale representations enhances fine-grained. We validate the performance through extensive experiments on three real-world datasets. The resutls compared with state-of-the-art methods demonstrate the superiority of the proposed model.

\end{abstract}
\section{Introduction}
With rapid urbanization, intelligent transportation systems (ITS) have become crucial components of modern smart cities\cite{song2014prediction,gutierrez2016international}. Using these systems, urban planners and administrators can obtain precise warnings regarding traffic congestion and public safety\cite{gong2020online,li2019sample}. For instance, the Halloween stampede accident in Itaewon, South Korea, resulted in 158 fatalities and 196 injuries. By deploying fine-grained crowd warning and prediction models, city managers can identify high-risk areas and proactively intervene before incidents occur\cite{li2024dual}. However, acquiring fine-grained flow data requires the deployment of a large number of devices throughout a city, leading to high operational costs in terms of procurement. To address these problems, fine-grained urban flow inference (FUFI) was recently proposed \cite{shao2022decoupled,wang2023fine}. As shown in Figure 1, the left image represents a coarse-grained flow map (32×32), while the right image displays a fine-grained flow (64×64). FUFI aims to accurately predict the distribution of fine-grained traffic flows by inferring them from the available coarse flow data. Moreover, it has a unique structural constraint: the total flow volume in the fine-grained regions strictly equals that of the corresponding coarse-grained regions.

Despite progress and achievements \cite{qu2022forecasting,chen2018neural} in addressing the FUFI problem\cite{Zhong2022FUFI}, most methods face a challenge in effectively and dynamically utilizing multi-scale geographical information. According to the third law of geography\cite{song2023geographically}, spatial correlation is not only related to distance but also to functionality\cite{gong2021missing,gong2023missingness}. Currently, few studies take this factor into account, often employing simple convolutional approaches that focus on local geographical distance correlation while overlooking global geographical functional correlation. In terms of time considerations, the correlation at each moment undergoes changes. For instance, in the morning, the foot traffic in supermarkets near homes and office buildings might be high, indicating strong correlation. However, as people leave the workplace in the evening, the correlation between these two supermarkets weakens. Nevertheless, capturing such dynamic relationships within this area is challenging and has also become a significant hurdle.

\begin{figure}[t]
\centering
\begin{minipage}[t]{0.7\linewidth}
    \includegraphics[width=\linewidth]{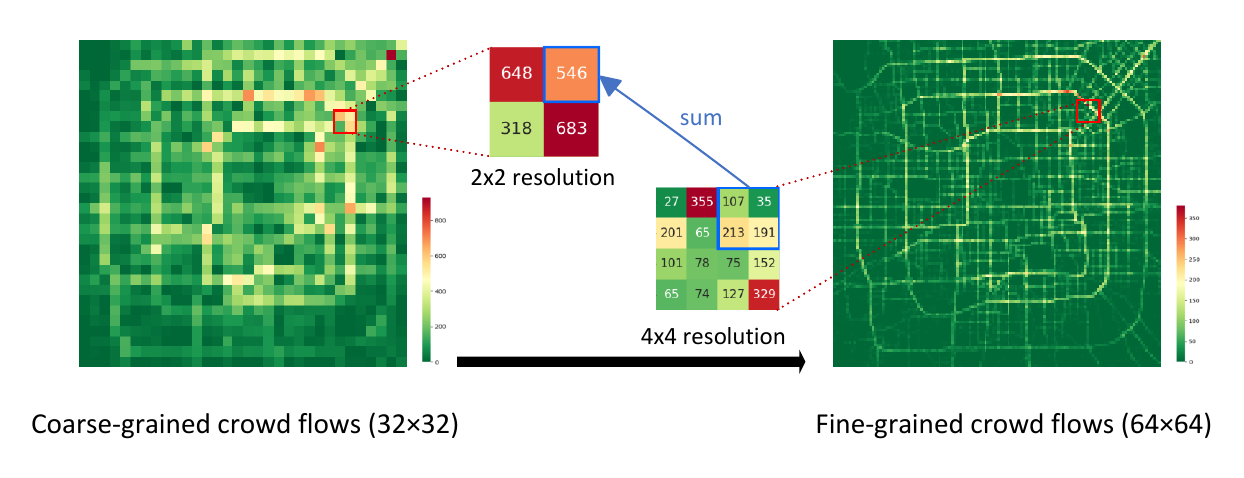}
     \caption{Fine-grained urban flow inference.}
\label{f1}
\end{minipage}%
\hfill%

\end{figure}
To address these issues, this paper proposes dynamic multi-scale geographic information\cite{liang2018geoman} fusion framework. Specifically, we introduce an encoder-decoder architecture and employ a two-stage training mechanism to accomplish fine-grained tasks. We design encoders at different scales to learn geographical distance correlation and geographical functional correlation. Simultaneously, we dynamically compare the encoders at different scales, dynamically selecting positive and negative samples at each moment, to obtain more meaningful spatiotemporal representations for each region. Furthermore, we propose private and interactive decoders to reconstruct the input, ensuring that multi-scale geographical information complements each other to faithfully restore high-quality fine-grained flow.

\section{RELATED WORK}
In this section, we investigate the current methods for fine-grained urban flow inference and single image super-resolution problems and reveal the differences between them.

{\bfseries Fine-Grained Urban Flow Inference.} The previous architecture heavily relies on empirically stacking deep neural networks. Liang et al. \cite{liang2019urbanfm} introduced the concept of structural constraints for the FUFI problem and designed an \begin{math}
    M^{2} 
\end{math}-normalization layer to satisfy these constraints. Since the proposal of neural ordinary differential equations (NODE), Zhou et al. \cite{zhou2020enhancing} 
discovered that NODE can be used as a core module to address the FUFI problem and proposed a more general FODE architecture. FODE effectively solves the numerical instability problem in previous methods without incurring additional memory costs. Qu et al. \cite{qu2022forecasting} proposed UrbanSTC to reduce the complexity and resource requirements of the model while still capturing important patterns in the data.

{\bfseries Single Image Super-Resolution.} Inspired by the superior performance of CNNs\cite{kim2016accurate,sun2008image}, various CNN models have been applied to SISR. 
Since Dong et al. \cite{dong2015image} first proposed an end-to-end mapping method using CNNs to restore high-resolution (HR) images from low-resolution (LR) images, various CNN-based architectures have been widely studied for Super Resolution (SR) research. However, deep networks are prone to model degradation during the training phase. Chen et al. \cite{chen2023dual} take into consideration information from two dimensions: image space and channels, for super-resolution.

However, there is a significant difference between the FUFI and image super-resolution tasks, specifically the unique structural constraint resent in FUFI.

\section{PROBLEM DEFINITION} In this section, we explain and formalize the FUFI (Fine-grained Urban Flow Inference) problem.

{\bfseries Definition 1 (Region)}:
According to the given longitude and latitude coordinates, we divide a city into a grid map with dimensions \begin{math}
    H\times W
\end{math}. Each grid element represents a region within the city.

{\bfseries Definition 2 (Superregion and Subregion)}:
As shown in Figure \ref{f1}, the scaling factor \begin{math} S=2\end{math} controls the resolution change between the coarse-grained grid map and the fine-grained grid map. Therefore, each larger coarse-grained grid is composed of \begin{math}
    2 \times 2
\end{math} smaller fine-grained grids. We define the aggregated the larger grid as a superregion and the smaller grids they consist of as subregions. In the FUFI problem, there exists a special structural constraint between the superregions and their corresponding subregions.

{\bfseries Definition 3 (Structural Constraint)}:
At timestamp \begin{math}
t\end{math}, the total sum of the flow in \begin{math}S\end{math}-by-\begin{math}S\end{math} subregions needs to be exactly equal to the flow in the corresponding superregion.
\begin{equation}
    x_{i,j}^{t}=\sum_{i',j'}y_{i',j'}^{t} 
\left \lfloor  \frac{i'}{S} \right \rfloor =i,\left \lfloor  \frac{j'}{S} \right \rfloor =j
,\end{equation}
where, \begin{math}
    i=1,2,...,H \end{math} and
    \begin{math} j=1,2,...,W
\end{math}.

{\bfseries Problem Definition (Fine-Grained Urban Flow Inference)}:
Given a coarse-grained map \begin{math}
    x^{t}
\end{math} at a timestamp \begin{math}
    t
\end{math} and a upscaling factor \begin{math}
    S\in \mathrm {Z}_{+} 
\end{math}, we can generate the corresponding fine-grained map \begin{math}
    y^{t}
\end{math} under the structural constraint.

\section{METHOD}
\begin{figure*}[t]
\begin{minipage}[t]{\linewidth}
    \includegraphics[width=\linewidth]{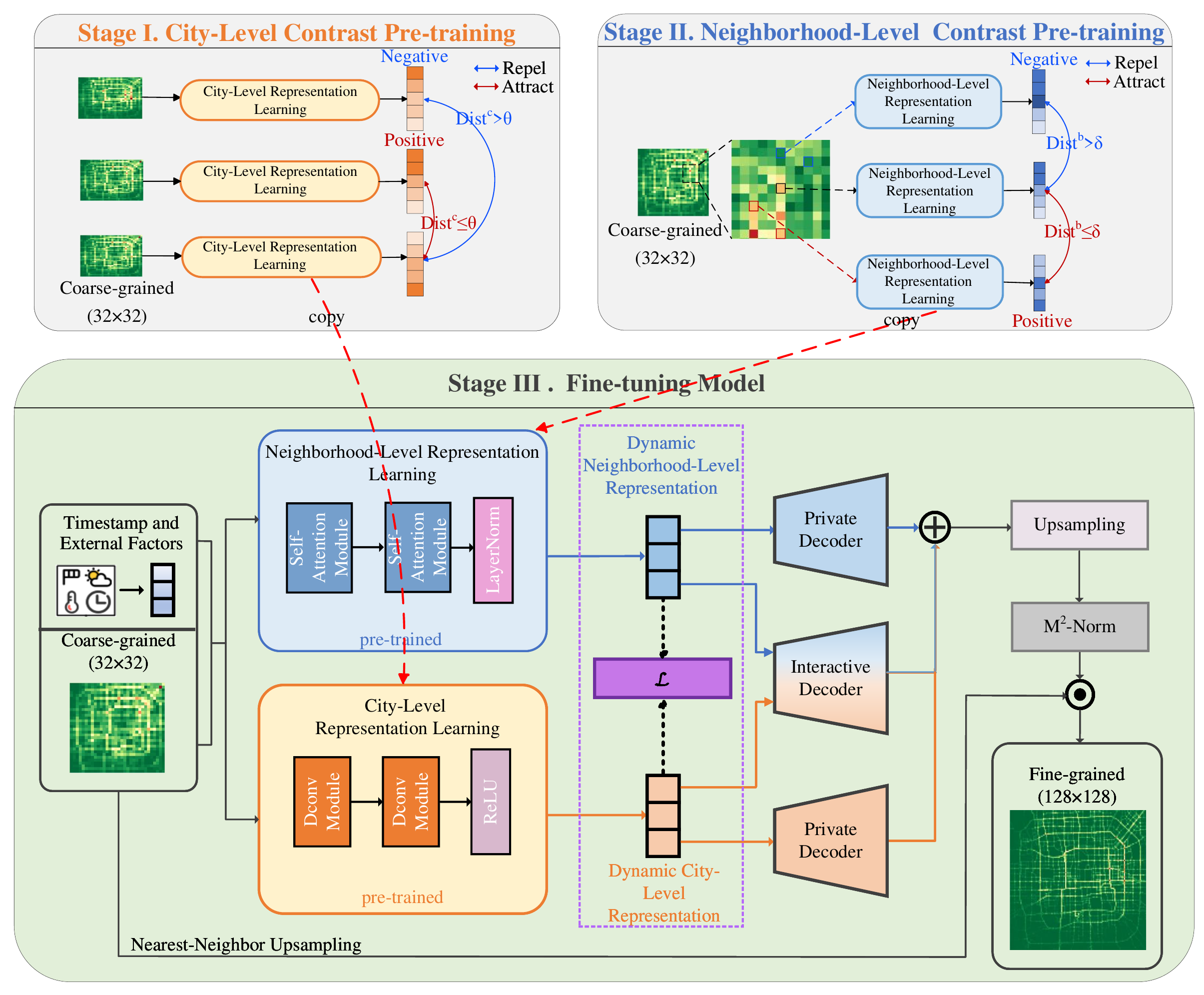}
    
    \caption{The overall structure of UrbanMSR consists of three main steps: Stage I and II focus on pre-training the neighborhood-level encoder and city-level encoder. Then, the pre-trained encoders are transferred to the model for fine-tuning (Stage III).
}
    \label{zhengti}
\end{minipage}%
    \hfill%

\end{figure*}

Figure \ref{zhengti} illustrates the framework of UrbanMSR, which is an encoder-decoder architecture. In this section, we list key components of the model.

\subsection{Dynamic Multi-Scale Representation Learning}
Two encoders are used to learn dynamic multi-scale representations.

{\bfseries Dynamic Positive and Negative Samples.} For neighborhood-level geographic representations, different regions exhibit varying relationships at different points in time. Therefore, we obtain positive and negative samples based on the Euclidean distance between regions at each moment:

\begin{equation}dist^{b}\left (x_{t,i,j},x_{t,i',j'}\right )=\sqrt{\left ( h_{t,i,j}^{b}-h_{t,i',j'}^{b}\right )^{2}  },
\end{equation}
where \begin{math} h_{t,i,j}^{b}\end{math} and \begin{math} h_{t,i',j'}^{b}\end{math} denotes the neighborhood-level representation at the same time for different regions.

\begin{equation} \begin{cases}
 P^{b}(x_{t,i,j})=x_{t,i',j'} & \text{ if } dist\left (x_{t,i,j},x_{t,i',j'}\right ) \le \delta \\
N^{b}(x_{t,i,j})=x_{t,i',j'}  & \text{ if } dist\left (x_{t,i,j},x_{t,i',j'}\right ) >\delta,
\end{cases}  
\end{equation}
where \begin{math}
   P^{b}(x_{t,i,j})
\end{math} are the positive samples of \begin{math}
    x_{t,i,j}
\end{math} and \begin{math}
    N^{b}(x_{t,i,j})
\end{math} are the negative samples of \begin{math}
   x_{t,i,j}
\end{math} at neighborhood-level.

For city-level representation, there is periodicity in all regions as a whole. Therefore, we use the mean square value of the Euclidean distance across all regions at different time points as the criterion for selecting positive and negative samples:
\begin{equation}dist^{c}\left (x_{t,i,j},x_{t',i,j}\right )=\sqrt{\frac{1}{HW} \sum_{i=1}^{H} \sum_{j=1}^{W}\left  ( h_{t,i,j}^{c}-h_{t',i,j}^{c}\right )^{2}  },
\end{equation}
where \begin{math} h_{t,i,j}^{c}\end{math} and \begin{math} h_{t',i,j}^{c}\end{math} denote the city-level representation of the same region at different time intervals.
\begin{equation} \begin{cases}
 P^{c}(x_{t,i,j})=x_{t',i,j} & \text{ if } dist\left (x_{t,i,j},x_{t',i,j}\right ) \le \theta\\
N^{c}(x_{t,i,j})=x_{t',i,j}  & \text{ if } dist\left (x_{t,i,j},x_{t',i,j}\right ) >\theta,
\end{cases}  
\end{equation}
where \begin{math}
   P^{c}(x_{t,i,j})
\end{math} are the positive samples of \begin{math}
    x_{t,i,j}
\end{math} and \begin{math}
    N^{c}(x_{t,i,j})
\end{math} are the negative samples of \begin{math}
   x_{t,i,j}
\end{math} at city-level.

{\bfseries Neighborhood-Level Contrast Pre-training.} A limitation of standard convolutions is their inability to effectively handle irregularly shaped neighboring regions. To address this deficiency, we choose deformable convolutions\cite{dai2017deformable} as the components of the encoder. Deformable convolutions introduce additional 2D offsets to the receptive field, allowing for more flexible and adaptive sampling of the input feature maps. For the representation \begin{math} z^{t}_{i,j}\end{math} of each region \begin{math}  \left ( i,j \right )  \end{math} at time \begin{math} t \end{math}, we have: 
 \begin{equation} z^{t}_{i,j}=\sum_{\left (i_{n},j_{n}\right )\in  R}w\left (i_{n},j_{n}  \right ) \cdot x^{t} \left ( i+i_{n}+\bigtriangleup i_{n},j+j_{n}+\bigtriangleup j_{n} \right ),   \end{equation}
where the regular grid  \begin{math} R \end{math} defines the receptive field size and dilation. For example, a kernel of the size \begin{math}3\times 3\end{math} with dilation 1. \begin{math}R=\left \{ (-1,-1),(-1,0),...,(0,1),(1,1) \right \} \end{math}. \begin{math} \left (i_{n},j_{n} \right)\end{math} denotes the locations within the grid \begin{math} R \end{math}. In deformable convolution, the regular grid \begin{math} R \end{math} is expanded by incorporating offsets \begin{math} \left \{ \left(\bigtriangleup i_{n},\bigtriangleup j_{n}\right)|n=1,...,N \right \}\end{math}, where \begin{math}N=|R|\end{math}.

Further processed use the ReLU activation function to obtain the neighborhood-level representation, 
\begin{equation}h^{b}=ReLu(z^{t} ),  \end{equation}
\begin{equation}E_{B}\left ( x^{t}  \right ) =h^{b}.   \end{equation}

Based on the previous section, we obtain Top-K positive and negative samples at the neighborhood-level for contrastive learning\cite{zhang2023spatio,zhang2023mask}. We prerain the neighborhood-level representation encoder using the contrastive loss. Our contrastive loss is expressed as:
\begin{equation} L_{b} =-log\frac{\sum_{k=1}^{K}sim\left (  x_{t,i,j},P^{b}(x_{t,i,j}) \right )_{k} }{\sum_{k=1 }^{K}sim\left (  x_{t,i,j},P^{b}(x_{t,i,j}) \right )_{k}+\sum_{k=1 }^{K}sim\left (  x_{t,i,j},N^{b}(x_{t,i,j})\right )_{k}},
\end{equation} 
where \begin{math} sim\left ( u,v \right )\end{math} is inner
product function between two representations.

{\bfseries City-Level Contrast Pre-training.} In terms of modeling spatial dependencies, previous work did not consider establishing long-range city-level spatial relationships. To capture global city spatial dependencies and inspired by the advantages of multi-head self-attention mechanism\cite{vaswani2017attention,devlin2018bert}. A multi-head self-attention (MHA) function can be defined as:
\begin{equation}
z^{c}=MHA\left ( w_{q}x_{p}^{t},w_{k}x_{p}^{t},w_{v}x_{p}^{t} \right ),
\end{equation}
where, \begin{math} w_{q},w_{k},w_{v}\in R^{HW\times d } \end{math} 
 and \begin{math} x_{p}^{t} \end{math} represents the representation of the original urban flow map \begin{math} x^{t} \end{math} after adding positional encoding.

The spatial encoder consists of multi-head self-attention layers and layernorm (LN) and residual connections in every block,
\begin{equation} h^{c} =LN\left ( x^{t}+z^{c}  \right ), 
\end{equation} 
\begin{equation}E_{c}\left ( x^{t}  \right ) =h^{c}.   \end{equation}

Similarly, based on the dynamic positive and negative samples obtained in the previous section for city-level representation, we prerain the city-level representation encoder using the contrastive loss.
\begin{equation} L_{c} =-log\frac{\sum_{k=1}^{K}sim\left (  x_{t,i,j},P^{c}(x_{t,i,j}) \right )_{k} }{\sum_{k=1 }^{K}sim\left (  x_{t,i,j},P^{c}(x_{t,i,j}) \right )_{k}+\sum_{k=1 }^{K}sim\left (  x_{t,i,j},N^{c}(x_{t,i,j})\right )_{k}}.
\end{equation}

\subsection{Fusion Of Multi-scale Representations.}To address the challenge of integrating different-scale geographical information, we utilize private decoders to obtain specific information for each scale.

{\bfseries Private Information Decoder.} The unique information at each scale needs to be preserved with a dedicated decoder featuring a distinctive structure. Therefore, the structure of the private information decoder can be designed based on the neighborhood-level and city-level encoders. The decoder adopts a network architecture similar to that of the encoder to ensure consistency and mutual adaptability.
\begin{equation}o^{b}=D_{B}\left ( h^{n}  \right ), 
\end{equation}
\begin{equation}o^{c}=D_{C}\left ( h^{c}  \right ).
\end{equation}

{\bfseries Interactive Information Decoder.} In the aspect of multi-scale information interaction, we aim to adopt a simple structure for information fusion. Therefore, the interactive information decoder blends the representations from the neighborhood-level and city-level using a straightforward convolution operation.
\begin{equation}o^{bc}=D_{BC}\left ( concat\left (h^{b},h^{c}\right )   \right ). 
\end{equation}

Then, the urban flow feature map \begin{math}
r^{t}\end{math} are combined scale-specific information and interactive information through weights. 
\begin{equation}
r^{t}=w_{1}o^{b}+w_{2}o^{c}+w_{3}o^{bc},
\end{equation}
where \begin{math}
w_{1}+w_{2}+w_{3}=1
\end{math}.

\subsection{Optimization}
Drawing inspiration from the distributional upsampling techniques employed in UrbanFM, we adopt a \begin{math}M^{2}-\end{math}Normalization approach that ensures the sum of subregions equals their corresponding superregion. This is utilized in generating fine-grained prediction values for loss calculation.

Our goal is to ensure diversity in multi-scale representations to enhance fine-grained accuracy. To achieve this, we introduce a discriminative loss function to alleviate redundancy. The formula for the feature differentiating loss is as follows:
\begin{equation}
L_{d} =-ReLu\left ( tanh\left (\frac{\alpha}{2HW}\sum_{i=1}^{H}\sum_{j=1}^{W}  (\left ( h_{i,j}^{b} \right ) ^{T}h_{i,j}^{c} +\left ( h_{i,j}^{c} \right ) ^{T}h_{i,j}^{c}  )\right )  \right ),
\end{equation}
where the features \begin{math}h^{b}\end{math} and \begin{math}h^{c}\end{math} represent the feature representations obtained after passing through the neighborhood-level encoder and city-level encoder.
Finally, we combine the conventional mean squared error (MSE) loss with a weighted feature differentiating loss as the model's overall loss to optimization the model:
\begin{equation}
    L=L_{MSE}  +\lambda L_{d},
\end{equation}
where, \begin{math}
    \lambda
\end{math} is the weight of the feature differentiating loss in the total loss.

\begin{table*}[t]
\centering
  \caption{The RMSE, MAE and MAPE on TaxiBJ, BikeNYC and DiDi-Xi'an. The best
results are bold and the second best are underlined.}
  \label{tab:commands}
  
  \begin{tabular}{cccc|ccc|ccc}
    \hline
     & \multicolumn{3}{c|}{TaxiBJ}& \multicolumn{3}{c|}{BikeNYC} & \multicolumn{3}{c}{DiDi-Xi’an} \\
     \cline{2-4}
     \cline{5-7}
     \cline{8-10}
    Methods &RMSE & MAE & MAPE&RMSE & MAE & MAPE&RMSE & MAE & MAPE\\
    \hline
     \texttt{MEAN} &23.684	&13.540& 4.887&2.012&	0.663	&0.393&10.782	&4.697	&2.416 \\
     
     \texttt{HA} &5.040	&2.384& 0.329&1.502&	0.364	&0.270&5.561	&1.192	&0.176 \\
     \hline
     \texttt{LapSRN} &5.260	&2.754&	0.512&	1.695&	0.555	&0.322&6.097	&1.850	&0.583 \\
    \texttt{VDSR} & 4.071 & 2.204	& 0.448&1.924&	0.636&	0.376 &6.469	&2.330	&1.038\\
    \texttt{SRResNet}& 3.976	&2.167&	0.453&1.824	&0.625&	0.370&5.608	&2.137&	1.015\\
    \texttt{IMDN}&4.095	&2.109	&0.352&1.950	&0.629&	0.371&5.579	&1.662	&0.557\\

    \texttt{DAT}&4.048	&2.069	&0.344 &1.531	&0.544&	0.317&5.293	&1.527	&0.519\\
    \hline
    \texttt{UrbanFM}& 3.951	&2.024	&0.304&0.711&	0.162&	0.070&5.091&	1.401&	0.375\\

     \texttt{UrbanPy}& 3.944&1.998&0.294&0.676&	0.147&	0.065&4.988&	1.361&	0.364\\
     
    \texttt{FODE}& 3.960& 2.030& 0.306&0.688	&0.149	&0.065&5.043	&1.380&	0.377\\
    
    \texttt{UrbanSTC}&\underline{3.918}	&\underline{1.988}	&	\underline{0.293}	&\underline{0.666}	&\underline{0.145}	&	\underline{0.064} &\underline{4.976}	&\underline{1.322}&\underline{0.352}	\\
    \hline
    \texttt{UrbanMSR}& \textbf{3.853}&	\textbf{1.952}&\textbf{0.286}&\textbf{0.654}	&\textbf{0.137}	&\textbf{0.057} &\textbf{4.886} &\textbf{1.271} &\textbf{0.328}\\
     
    \begin{math}
        \Delta
    \end{math} & +1.65\% &	+1.81\% & +2.38\% & +1.80\% & +5.51\% & +10.90\% & +1.80\% & +3.85\% & +6.81\%\\
     
    \hline
    
  \end{tabular}
  \label{jieguo}
\end{table*}

\section{EXPERIMENTS}
\subsection{Experimental Settings}
We assess the effectiveness of our model and baselines on three real-world urban flow datasets. To ensure the feasibility of the experiments, we selected three datasets: a large dataset from Beijing\cite{zhang2017deep}, as well as two smaller datasets from New York and Xi'an. We compare the proposed method UrbanSTC with 11 baseline methods. Mean Partition (MEAN) and Historical Average (HA) are heuristic methods. LapSRN \cite{lai2017deep}, VDSR \cite{kim2016accurate}, SRResNet \cite{ledig2017photo}, IMDN \cite{hui2019lightweight} and DAT \cite{chen2023dual} are image super-resolution methods. UrbanFM \cite{liang2019urbanfm}, UrbanPy \cite{ouyang2020fine}, FODE \cite{zhou2020enhancing}, UrbanSTC \cite{qu2022forecasting} are FUFI methods. We assess the performance of various methods using three commonly used metrics: Root Mean Squared Error (RMSE), Mean Absolute Error (MAE), and Mean Absolute Percentage Error (MAPE).

\subsection{Results On Datasets}

We evaluated the performance of our model and baseline methods on different datasets, and the results are shown in Table \ref{jieguo}. Our model outperformed all competing methods across the various datasets. Specifically, when compared to the current state-of-the-art FUFI method, our model achieved improvements of 1.65\% in RMSE, 1.81\% in MAE, and 2.38\% in MAPE on the TaxiBJ dataset. Our model also demonstrated improvements of 1.80\% in RMSE, 5.51\% in MAE, and 10.90\% in MAPE on the BikeNYC dataset and  1.80\% in RMSE, 3.85\% in MAE, and 6.81\% in MAPE on the DiDi-Xi'an dataset. The above results indicate that our motivation is justified. Learning different multi-scale geographic information can lead to better representation learning and improved performance in the FUFI task.

\subsection{Ablation Analysis}
In this section, we used TaxiBJ to analyze the roles of each model component in the model. “neighborhood-level” represents the deformable convolution encoder–decoder branch that focuses on capturing information from geographic data, "city-level" represents the multi-head self-attention encoder–decoder branch that focuses on capturing global geographic information. By observing Table \ref{xr2jg}, we found that the original model performs better than any individual single branch, which demonstrates that the model has learned multi-scale information and balanced multiple sources of information. Notably, the final fusion model outperforms each individual branch, indicating that the semantic information learned from each branch is unique and complementary, contributing to the inference of fine-grained urban flow.

\begin{table}[t]
\centering
  \caption{Ablation study of dynamic encoders and external factors. We report the strategies used in different models on TaxiBJ dataset’s results.}
  \label{tab:freq}
  \resizebox{0.7\linewidth}{!}{
  \begin{tabular}{cc|ccc}
    \hline
      \multicolumn{2}{c|}{}& \multicolumn{3}{c}{TaxiBJ}  \\
     
     \cline{3-5}
     neighborhood-level & city-level &RMSE & MAE & MAPE\\
     \hline
   
    \checkmark & & 3.929&	1.985	&0.290\\
   
     	&\checkmark&3.868	&1.970	&0.297\\

     \checkmark	&\checkmark &\textbf{3.853}	&\textbf{1.952}	&\textbf{0.286}\\
   
  \hline
\end{tabular}}
\label{xr2jg}
\end{table}

\subsection{End-to-end and two-stage Comparison.}
To verify the role of the dynamic pre-trained encoders\cite{yang2021self}, including a neigh-
borhood-level encoder and a city-level encoder, we conducted experiments on the TaxiBJ and BikeNYC datasets as shown in Table \ref{xr1jg}. We can clearly observe that the results from the two-stage experimental process are superior to the end-to-end training process. The two-stage approach, employing a pre-trained encoder, was utilized to learn dynamic information from different regions, thereby gaining a better understanding of each region and enhancing the granularity accuracy in each region.

\begin{table}[t]
\centering
  \caption{End-to-End and two-stage comparison.}
  \label{tab:freq}
  \resizebox{0.7\linewidth}{!}{
  \begin{tabular}{cccc|ccc}
    \hline
     & \multicolumn{3}{c|}{TaxiBJ}& \multicolumn{3}{c}{BikeNYC}  \\
     \cline{2-4}
     \cline{5-7}
    Methods &RMSE & MAE & MAPE&RMSE & MAE & MAPE\\
    \hline
    End-to-End&3.884&	1.965	&0.288&0.686&	0.140	&0.058\\
    \hline
    Two-stage & \textbf{3.853}	&\textbf{1.952}	&\textbf{0.286}&\textbf{0.654}&	\textbf{0.137}&	\textbf{0.057}\\
   
  \hline
\end{tabular}}
\label{xr1jg}
\end{table}

\section{CONCLUSION}
In this research, we introduced UrbanMSR, which includes specially designed encoders for dynamically capturing subtle differences in multi-scale geographic information. We employ feature-guided loss functions at different scales to ensure diversity of representations. In addition, the proprietary and shared decoders are a good blend of multi-scale representation. Through extensive experiments on three different real-world datasets, we have demonstrated the power of UrbanMSR in the field of complex fine-grained stream inference.

\subsubsection{\ackname} This work was supported in part by the National Natural Science Foundation of China under Grant 62202270, in part by the Natural Science Foundation of Shandong Province, China, under Grant ZR2021QF034, in part by the Shandong Excellent Young Scientists Fund (Oversea) under Grant 2022HWYQ-044, and in part by the Taishan Scholar Project of Shandong Province under Grant tsqn202306066.


%
%
%
\bibliographystyle{splncs04}
\bibliography{myarticle}
%





\end{document}